# MULTIPLE CONFIGURATIONS FOR PUNCTURING ROBOT POSITIONING

Omar Abdelaziz[134], Minzhou Luo[23], Guanwu Jiang[13] and Saixuan Chen[13]

[1]University of Science and Technology of China, Hefei, 230026, China

[2]Key Laboratory of special robot technology of Jiangsu Province, Hohai University, Changzhou, 213000, China

[3]Institute of Intelligent Manufacturing Technology, Jiangsu Industrial Technology Research Institute, Nanjing, 211800, China

[4] Egyptian Russian University, Egypt

## ABSTRACT

*The paper presents the Inverse Kinematics (IK) close form derivation steps using combination of analytical and geometric techniques for the UR robot.  The innovative application of this work is used in the precise positioning of puncture robotics system. The end effector is a puncture needle guide tube, which needs precise positioning over the puncture insertion point. The IK closed form solutions bring out maximum 8 solutions represents 8 different robot joints configurations. These multiple solutions are helpful in the puncture robotics system, it allow doctors to choose the most suitable configuration during the operation. Therefore the workspace becomes more adequate for the coexistence of human and robot. Moreover IK closed form solutions are more precise in positioning for medical puncture surgery compared to other numerical methods. We include a performance evaluation for both of the IK obtained by the closed form solution and by a numerical method.*

## KEYWORDS

*MIS robotics system, inverse kinematics, puncture positioning, UR robot*

## 1. INTRODUCTION

There are growing demands in the usage of robots nowadays and there has been a lot of research carried out in robotics fields. The robotics applications can be seen in transportation, industry, medical, defence, service, etc. Robotics research has moved rapidly in the last few years toward the field of medical and health care. Today, medical robotics considered a reliable solution [1]. Strength points for robots over human in medical applications [2] i.e. achieving minimal invasiveness, geometric accuracy, no fatigue or inattention, stable, repeatability, radiation insensitive, and  integration capability of numerical & sensor data. In literature, there are several early attempts to categorize and classifies the applications of robotics systems in medical and health care. They are mainly three classifications as in [3], [4]: macro-robotics, micro-robotics and bio-robotics. Macro- robotics includes the development of assistive and rehabilitation robots as well as new, more powerful tools and techniques for surgery. Micro-robotics could greatly contribute to the field of minimally invasive surgery as well as to the development of a new generation of tools for conventional surgery. Bio-robotics based on modelling and simulating biological systems. Several examples of medical robotics systems currently exists such as robotic orthopaedic surgery, robotically assisted percutaneous therapy, minimally invasive robotic surgery, rehabilitation and assistive devices, and laboratory [5], [6]. Our interest in this paper goes to the Minimally-Invasive Surgical (MIS) robotics system. Surgical robotics system in general can be thought of as smart surgical tools that enable human surgeons to treat individual patients with improved efficacy, greater safety, and less morbidity than would otherwise be possible. It is an integration of a number of modern and complex high technologies that let doctors (surgeons), through the robotics system, can perform surgical operations without touching patients. MIS robot system is a combination of medical image processing technology and the operation of the mechanical arm to perform puncture surgery on the patient [7]. It can be divided in to two branches; an assistive capability, i.e. tool positioning, and an active capability, i.e. for conducting transurethral resection of the prostate and arthroplasty (joint repair). MIS is an operation performed using specialized instruments designed to fit into the body through several tiny punctures instead of one large incision [8].

This paper presents a solution for an adequate workspace for surgical robotics system by using a 6 DOF robot with obtainable closed form IK. We mean by Adequate Workspace; that the workspace is convenient for human to coexist with robot either in a collaborative manner or a non-collaborative manner. Where by solving IK we can get maximum 8 solutions, which allow doctors to choose the most suitable configuration during the operation. The suitable configuration could be the one which give doctors more space, mobility, visibility, etc. For solving the problem of IK, the kinematics model needs to be determined. The analysis of kinematic can solve the problem of mapping the relation of the joints angle and the end effector pose. This is the theoretical basis of motion control and one of the most important steps to realize close-loop control accurately [9]. This paper work is applied on the UR3 robot, one of the popular cooperative robots in the market [10]. We present the robot forward kinematics and give review on the available methods for solving the IK problem. A complete IK closed form solution obtained by mixing the geometric and analytical techniques is presented. The solution for the joint1 ($q_1$), joint5 ($q_5$), joint6 ($q_6$), and the summation of joints 2, 3 and4 ( $q_{234} = q_2 + q_3 + q_4$) obtained analytically. Then the values for $q_2, q_3, q_4$ obtained geometrically. The closed form solution resulting to maximum 8 solutions for a given end-effector pose in task space. However, for some other serial robots it could be infeasible to obtain the closed form solution. For this reason, we present the Steps for obtaining IK using The Gauss - Newton iterative method. We include in this work performance evaluation between IK obtained from the closed form solution and numerical solution using The Gauss - Newton iterative method.

The paper is organized in this manner. Section 1 describes the puncture robotic system. Section 2 describes the forward kinematics. Section 3 shows the different approaches of IK. Section 4 presents the Experimental Results and evaluation. Finally, Section 5 gives conclusions.

## 2. MIS ROBOTICS SYSTEM

The puncture robotics system is highly demanded in medicine field due to its advantages such as achieving minimal invasiveness, accuracy, efficiency, and stability. The system is based on Computerized Tomography (CT) image processing and its three-dimensional reconstruction that guide a 6 DOF robot toward precise positioning for MIS. The system is illustrated in Figure 1; its operation is explained in the following steps. First step the patient is required to do CT scan. Then a three-dimensional model is reconstructed from the CT scan image of the patient. Through the three-dimensional model doctor can observe and diagnose the disease. Second step the doctor determines the position of the puncture target point and insertion point coordinates. The connection between the target point and insertion point is the green line in Figure 1; this line represents the puncture route. The route must avoid the patient's bones, blood vessels, and other organs. The accurate puncture route determines the quality of the puncturing operation. Then the target point and insertion point coordinates are sent to the robot through the data processing computer. The robot end-effector is a needle guide tube shown in Figure 2, accurately positions the needle to the patient's skin according to the puncturing route direction. Finally, the needle insertion is done manually by the doctor.

### 2.1 Experimental Setup

We developed a computer Graphical User Interface GUI visualization system for doctors based on the CT image. For simulating the system, we designed a 3D model typify the chest. The experimental setup is shown in Figure 2. The marks on the 3D model in black and white colours are for image processing purposes. Next step we do CT scan for the 3D model. We insert these images in our developed GUI system. The doctor can choose the insert point and target point through the GUI system, the full system include the vision implementation on an animal experiment is described in details in our previous work in [11]. Then the robot is commanded to position the end effector at the insertion point. This step is the concern of this paper work. As by conducting the work in this paper, we are able to offer the doctor multiple robot configurations to choose from them the most adequate one for him. The coming sections explain the steps to obtain the IK solutions.

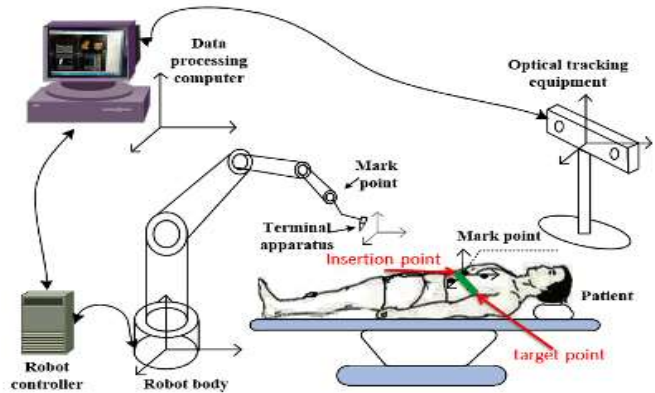

**Figure1.** Puncture robotics system, consists of a 6Dof robot manipulator and the computer for image processing and robot control

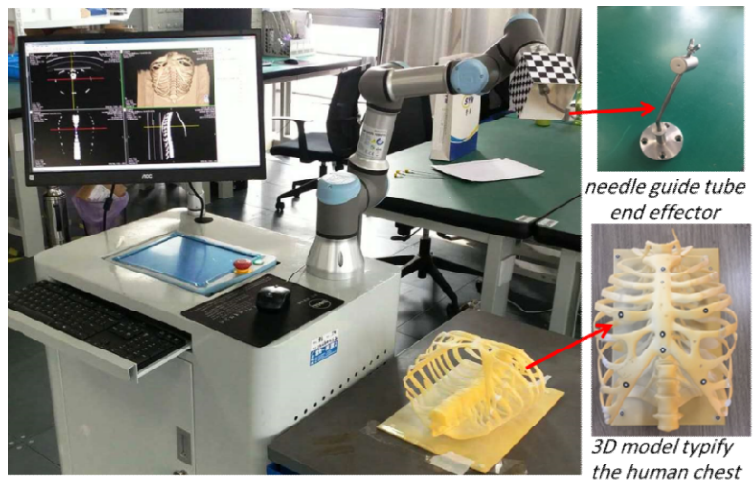

**Figure2.** Experimental setup, showing the GUI, 3D chest model and needle guide tube end-effector

## 3. FORWARD KINEMATICS

Kinematic modelling based on the Denavit-Hartenberg (D-H) parameters [12]. The DH parameters are minimal representation of the robot kinematics, which consists of 4 parameters ( $a_i, d_i, \alpha_i, \theta_i$ ). The UR robot design and its free body diagram (FBD) with links reference frames shown in Figure 3. The DH parameters are shown in Table 1Table 1. These 4 parameters can define the transformation from coordinate frame link $\{i+1\}$ to frame $\{i\}$ in terms of four elementary sequences of rotations and translations [13], [14], first a rotation around z axis by $\theta_i$ then a translation about the z axis by $d_i$ followed by translation about x axis by $a_i$ and then rotation around x axis by angle $\alpha_i$ . This transformation can be represented as a matrix A

$$^{i-1}_{i}A(a_i, \theta_i, d_i, \alpha_i) = Rz(\theta_i)Tz(d_i)Tx(a_i)Rx(\alpha_i) \tag{1}$$

$$^{i-1}_{i}A = \begin{bmatrix} C\theta_i & -S\theta_i & 0 & 0 \\ S\theta_i & C\theta_i & 0 & 0 \\ 0 & 0 & 1 & 0 \\ 0 & 0 & 0 & 1 \end{bmatrix} \begin{bmatrix} 1 & 0 & 0 & 0 \\ 0 & 1 & 0 & 0 \\ 0 & 0 & 1 & d_i \\ 0 & 0 & 0 & 1 \end{bmatrix} \begin{bmatrix} 1 & 0 & 0 & a_i \\ 0 & 1 & 0 & 0 \\ 0 & 0 & 1 & 0 \\ 0 & 0 & 0 & 1 \end{bmatrix} \begin{bmatrix} 1 & 0 & 0 & 0 \\ 0 & C\alpha_i & -S\alpha_i & 0 \\ 0 & S\alpha_i & C\alpha_i & 0 \\ 0 & 0 & 0 & 1 \end{bmatrix}$$

$$\tag{2}$$

$$^{i-1}_{i}A = \begin{bmatrix} C\theta_i & -C\alpha_i S\theta_i & S\alpha_i S\theta_i & a_i C\theta_i \\ S\theta_i & C\alpha_i C\theta_i & -S\alpha_i C\theta_i & a_i S\theta_i \\ 0 & S\alpha_i & C\alpha_i & d_i \\ 0 & 0 & 0 & 1 \end{bmatrix}$$

Then the transformation matrix for the End Effector with respect to the base frame is obtained by multiplying the transformation matrix for each join gradually from joint 1 to joint n, in our case (n = 6). Transformation matrix $^{0}_{6}A$ as function of joints position is

$$^{0}_{6}A(q_1, \ldots, q_6) = \prod_{i=1}^{6} {}^{i-1}_{i}A \equiv \begin{bmatrix} R_{3\times3} & P_{3\times1} \\ 0 & 1 \end{bmatrix} \tag{3}$$

From these formulas above, we should have a forward kinematics representation of the robot.

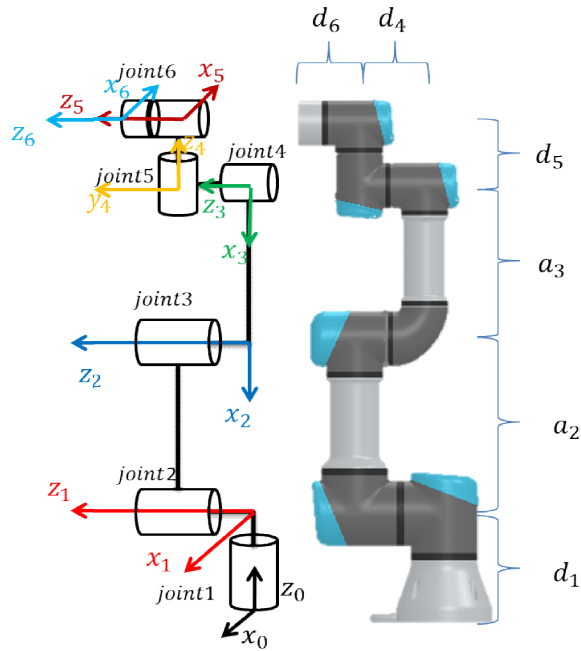

Figure 3. UR Robot FBD with links reference frames

Table 1. UR Robot DH parameters

| joint | $\theta_i$(rad) | $a_i$ (m) | $d_i$ (m) | $\alpha_i$ (rad) |
|---|---|---|---|---|
| 1 | $\theta_1$ | 0 | $d_1$ | $\pi/2$ |
| 2 | $\theta_2$ | $a_2$ | 0 | 0 |
| 3 | $\theta_3$ | $a_3$ | 0 | 0 |
| 4 | $\theta_4$ | 0 | $d_4$ | $\pi/2$ |
| 5 | $\theta_5$ | 0 | $d_5$ | $-\pi/2$ |
| 6 | $\theta_6$ | 0 | 0 | 0 |

## 4. INVERSE KINEMATICS

The problem of inverse kinematics can be stated as follow: what joints variates $q_i (i = 1 \rightarrow 6)$ for a desired End Effector position and orientation $(n_x, n_y, \ldots p_y, p_z)$. In the field of engineering application, inverse kinematics is more important than forward kinematics. It's the base of motion planning and trajectory control. Obtaining the closed form solution analytically is not feasible for many cases. Thus the need for obtaining the inverse kinematics numerically is significant. For the Closed form solution, there are two sufficient conditions in order to get analytical closed-form solution [13], [15], [16]. First if there are 3 adjacent axes of the joints intersect at one point, such as in manipulators with spherical wrist. Second if there are 3 adjacent axes of the joints parallel to each other. Fortunately The UR3 robot meets the second condition. So, the closed-form solution is feasible. Next, the closed form solution is derived, and then we show an alternative method using The Gauss - Newton iterative algorithm.

### 4.1 Closed Form Solution

The problem of the inverse kinematics that it does not have unique solution and this could be a problem in trajectory planning. The UR3 robot has 8 solutions for the end effector. We say that the shoulder $q_1$ have two configurations (Right-Left), the elbow $q_2$ and $q_3$ have two configurations (Up-Down) then the wrist $q_4$ has two configurations (Folded-Unfolded). Here we present complete solution using a mix between analytical and geometric techniques. We start by getting the analytical solution for $q_1, q_5, q_6$ and $q_{234}$ . The desired end effector pose is given in this form

$$
{}^0_6T_d = \begin{bmatrix} n_x & o_x & a_x & p_x \\ n_y & o_y & a_y & p_y \\ n_z & a_z & a_z & p_z \\ 0 & 0 & 0 & 1 \end{bmatrix} \tag{4}
$$

Where the desired orientation and position is

$$R_{d_{3\times3}} = \begin{bmatrix} n_x & o_x & a_x \\ n_y & o_y & a_y \\ n_z & o_z & a_z \end{bmatrix}, P_{d_{3\times1}} = \begin{bmatrix} p_x \\ p_y \\ p_z \end{bmatrix} \tag{5}$$

From the kinematics transformation matrix in equation (3) we infer that

$$^0_6A(q_1, \ldots, q_6) = {}^0_6T_d \tag{6}$$

Where;

$$\tag{7}$$

$$^1_6A(q_2, \ldots, q_6) = {}^1_2A\,^2_3A\,^3_4A\,^4_5A\,^5_6A$$

$$^1_6T(q_1) = ({}^0_1A)^{-1} * {}^0_6T_d \tag{8}$$

Then we do some manipulation to find the solutions. First using the equations (7) and (8), which are equivalent, to find $q_1, q_5$ and $q_6$.

$q_1$ is found from comparing 3rd and 4th Columns for the 3rd Row of $^1_6A$ and $^1_6T$ , as

$$q_1 = \text{atan2}\left(d_4, \pm\sqrt{\left(d_6a_y - p_y\right)^2 + (p_x - d_6a_x)^2 - d_4{}^2}\right) \\ - \text{atan2}\left(d_6a_y - p_y, p_x - d_6a_x\right) \tag{9}$$

There are two solutions for $q_1$ the (Right and Left) shoulder configurations. We notate the value obtained from equation (9) using the (+ve) sign with $q_{1,1}$ and the one from the (–ve) sign with $q_{1,2}$.

$q_5$ and $q_6$ can found From comparing 1st and 2nd Columns for the 3rd row of $^1_6A$ and $^1_6T$ obtained from the equations (7) and (8) , where

$$q_5 = \text{atan2}\left(\pm\sqrt{\left(n_xS_1 - n_yC_1\right)^2 + \left(o_xS_1 - o_yC_1\right)^2}, a_xS_1 - a_yC_1\right) \tag{10}$$

$q_5$ have two solutions, the one obtained from using the +ve sign in equation (10) and substituting with $q_{1,1}$ notated as $q_{5,1}$. The one obtained from using the -ve sign in equation (10) and substituting with $q_{1,2}$ notated as $q_{5,2}$. Later we will get another two solutions for $q_5$ from the wrist configurations, so we do it step by step.

$$q_6 = \text{atan2}\left(\frac{-o_xS_1 - o_yC_1}{S_5}, \frac{n_xS_1 - n_yC_1}{S_5}\right) \tag{11}$$

Also there are two solutions for $q_6$. First is $q_{6,1}$, which obtained from substituting in equation (11) by $q_{1,1}$ and $q_{5,1}$. Second is $q_{6,2}$, which obtained from substituting in equation (11)( by $q_{1,2}$ and $q_{5,2}$. The full derivation steps for $q_1, q_5$ and $q_6$ are presented in our previous work in [15].

Next, $q_{234} = (q_2 + q_3 + q_4)$ is obtained from coming two new equations (12) and (13).

$$_4^0A(q_1, \dots, q_4) = {_1^0A}{_2^1A}{_3^2A}{_4^3A} \qquad (12)$$

$$_4^0T(q_5, q_6) = {_6^0T_d}\left(_6^5A\right)^{-1}\left(_5^4A\right)^{-1} \qquad (13)$$

These two equations are equivalent. By observing the equations we found that the 3rd row of the 1st column in $_4^0A$ is

$$_4^0A(3,1) = \sin(q_{234}) \qquad (14)$$

And the 3rd row of the 3rd column in $_4^0A$ is

$$_4^0A(3,3) = -\cos(q_{234}) \qquad (15)$$

And there equivalences in in $_4^0T$ are;

$$\qquad (16)$$

$$_4^0T(3,1) = n_z C_5 C_6 - a_z S_5 - o_z C_5 S_6$$

$$_4^0T(3,3) = -o_z C_6 - n_z S_6 \qquad (17)$$

Where $C_6 = \cos(q_6)$, and $S_6 = \sin(q_6)$, similarly $C_i = \cos(q_i)$, and $S_i = \sin(q_i)$. Hence,

$$q_{234} = \text{atan2}(_4^0A(3,1), -_4^0A(3,3)) = \text{atan2}(_4^0T(3,1), -_4^0T(3,3)) \qquad (18)$$

$$q_{234} = \text{atan2}(n_z C_5 C_6 - a_z S_5 - o_z C_5 S_6, o_z C_6 + n_z S_6) \qquad (19)$$

Consequently, we have two solutions for $q_{234}$. The one obtained by Substituting $q_{5,1}$ and $q_{6,1}$ in equation (19) is $q_{234,1}$, and $q_{234,2}$ when substituting by $q_{5,2}$ and $q_{6,2}$.

Next step is to get $q_2$ and $q_3$ from the geometry of the robot. In Figure 3, the robot FBD, it is shown that joints 2, 3 and 4 are parallel, and it can be seen that reference frame 4 is shifted from joint 4 by the distance $d_4$. And by excluding joint 5 and 6, the problem is reduced into a planer robot to be illustrated as shown in Figure 4. For solving the problem, the position at joint 4 is required. We can get the position at the origin of reference frame 4 from $_4^0T$ in equation (13), as the position $p_4$ is the 4th column with respect to the base frame. We found the values for $p_4 = [p_{x4} \quad p_{y4} \quad p_{z4}]^T$ are;

$$p_{x4} = p_x - a_x d_6 + d_5 o_x C_6 + d_5 n_x S_6 \tag{20}$$

$$p_{y4} = p_y - a_y d_6 + d_5 o_y C_6 + d_5 n_y S_6 \tag{21}$$

$$p_{z4} = p_z - a_z d_6 + d_5 o_z C_6 + d_5 n_z S_6 \tag{22}$$

$p_4$ has two solutions for each $q_6$. So we get $p_{4,1} = [p_{x4,1} \quad p_{y4,1} \quad p_{z4,1}]^T$ from substituting in equations (20) (21) (22) by $q_{6,1}$, respectively for $p_{4,2}$ substituting by $q_{6,2}$ .

Then from the robot geometry top view shown in

Figure4(b) we can extract some other date. First the projection length ($r_1$) of links 2 and 3, can be derived as follow

$$l = \sqrt{p_{x4}^2 + p_{y4}^2} \tag{23}$$

$$r_1 = \sqrt{l^2 - d_4^2} \tag{24}$$

By using the $r_1$ value, we can estimate the other parameters from the side view in Figure 4(a), where the height in the z direction $r_2$ is

$$r_2 = p_{z4} - d_1 \tag{25}$$

And

$$r_3 = \sqrt{r_1^2 + r_2^2} \tag{26}$$

Here also have two solutions for each of ($l$, $r_1 r_2$ and $r_3$) according to which $p_4$ is used. Where ($l_1$, $r_{1,1} r_{2,1} r_{3,1}$) obtained by $p_{4,1}$, and ($l_2$, $r_{1,2} r_{2,2} r_{3,2}$) obtained by $p_{4,2}$.

Also q3 can be calculated from Figure 4(a), the side view. As we can get these two relations

$$r_1 = a_2 C_2 + a_3 C_{23} \tag{27}$$

$$r_2 = a_2 S_2 + a_3 S_{23} \tag{28}$$

Then By squaring and summing the two equations (27) and (28) and applying the sum and difference trigonometric we get

$$r_1^2 + r_2^2 - (a_2^2 + a_3^2) = 2a_2 a_3(C_{23}C_2 + S_{23}S_2) = 2a_2 a_3 \cos(q_{23} - q_2) \tag{29}$$

Since;

$$\cos(q_{23} - q_2) = \cos(q_3) \equiv C_3 \tag{30}$$

Then

$$C_3 = \frac{r_1{}^2 + r_2{}^2 - (a_2{}^2 + a_3{}^2)}{2a_2a_3} \tag{31}$$

So

$$\sin(q_3) = \pm\sqrt{1 - C_3{}^2} \tag{32}$$

Therefore

$$q_3 = \text{atan2}\left(\pm\sqrt{1 - C_3{}^2}, C_3\right) \tag{33}$$

There are four solutions for $q_3$ depends on which ($l, r_1 r_2$ and $r_3$) is used. Where $q_{3,1}$ obtained from +ve solution in equation (33) and $q_{3,2}$ obtained from -ve solution substituting by ($l_1, r_{1,1} r_{2,1} r_{3,1}$). Similarly, $q_{3,3}$ obtained from +ve solution in equation (33) and $q_{3,4}$ obtained from -ve solution substituting by ($l_2, r_{1,2} r_{2,2} r_{3,2}$). For now we have solutions for $q_3$

Next, to compute $q_2$ we need to extract $\alpha$ and $\beta$, shown in Figure 4(a). $\alpha$ and $\beta$ can be calculated as follow

$$\alpha = \text{atan } 2(r_2, r_1) \tag{34}$$

$$\beta = \text{atan2}(a_3 S_3 a_2 + a_3 C_3) \tag{35}$$

Where $\alpha$ have two solutions based on which $r_2, r_1$ are used, where ($l_1, r_{1,1} r_{2,1} r_{3,1} \rightarrow \alpha_1$) and ($l_2, r_{1,2} r_{2,2} r_{3,2} \rightarrow \alpha_2$). In the other side, $\beta$ have four solutions based on which $q_3$ is used, where ( $q_{3,1} \rightarrow \beta_1$ ), ( $q_{3,2} \rightarrow \beta_2$ ), ( $q_{3,3} \rightarrow \beta_3$ ) and ( $q_{3,4} \rightarrow \beta_4$ ). From the same figure it is shown that

$$q_2 = \alpha - \beta \tag{36}$$

$q_2$ have four solutions , where ( $q_{2,1} = \alpha_1 - \beta_1$ ), ( $q_{2,2} = \alpha_1 - \beta_2$ ), ( $q_{2,3} = \alpha_2 - \beta_1$ ) and ( $q_{2,4} = \alpha_2 - \beta_2$ ). For the four solutions of $q_2$ and $q_3$, a set of two of them ( $q_{2,1}, q_{2,2}, q_{3,1} q_{3,2}$ ) were obtained at the right shoulder configuration ($q_{1,1}$), and the other set ( $q_{2,3}, q_{2,4}, q_{3,3} q_{3,4}$ ) are obtained at the left configuration ($q_{1,2}$). The set from the right configuration have no problem. The left configuration have no problem with $q_5$ and $q_6$ because they were calculated based on the configuration of $q_1$, but the problem is in the other 3 joints as to calculate $q_4$. Therefore, for the left configuration at $q_{1,2}$, $q_2$ must be subtracting from pi, equivalent to mirroring around z-axis and q3 must be negate by take the negative of it, equivalent to mirroring around x-axis. Hence, the new values at the left configuration are

$$q_{2,3} = \pi - q_{2,3}$$
$$q_{2,4} = \pi - q_{2,4}$$
$$q_{3,3} = -q_{3,3}$$
$$q_{3,4} = -q_{3,4}$$

(37)

By using these new values for $q_2$ and $q_3$, the four solutions of $q_4$ can be calculated from

$$q_4 = q_{234} - q_3 - q_2$$

(38)

At this level, four different solutions of the inverse kinematics are obtained. The solutions have two configurations for the shoulders (Right- Left) and two for the elbows (Up- Down). There are still two solutions for the wrist configuration (Folded-Unfolded) as shown in Figure 5. We found that $q_6$ in equation (11) is dependent on $q_1$ and $q_5$; and since $q_1$ is fixed in both wrist configuration, then $q_5$ is the only dependent variable. Similarly for $q_{234}$ in equation (19) is dependent on $q_6$ and $q_5$. Therefore by flipping $q_5$, it can leads to new $q_6$, consequently for new $q_{234}$. The two new flipped values for $q_5$ are ($q_{5,3} = -q_{5,1}$ at ($q_{1,1}$), and ( $q_{5,4} = -q_{5,2}$ at ($q_{1,2}$). Then By following the same steps from equation (19) to (38) using the new values, it will end up with new four solutions. The eight solutions are arranged and shown in Table 2.

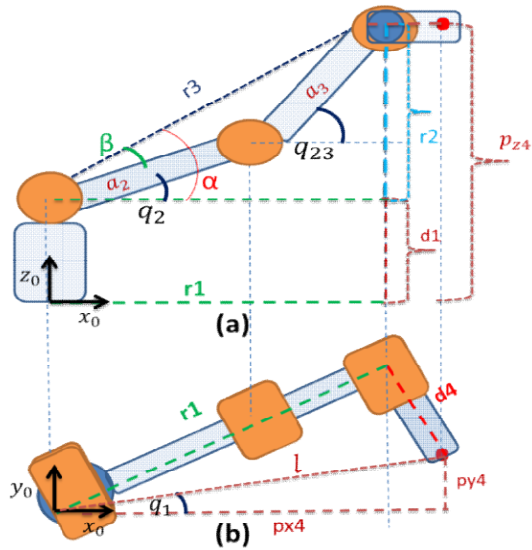

Figure **4.** Robot geometry for joints 2, 3 and 4 FBD (a) side view, (b) top view

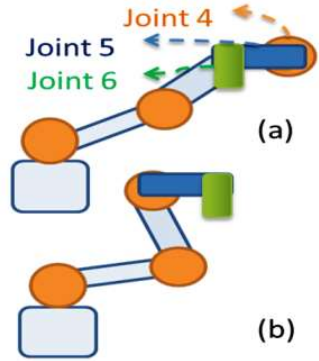

Figure 5. Joints 4, 5 and 6 at two different configurations Folded (a) and Unfolded (b)

Table 2. The inverse kinematics solutions for 8 configurations

| Configuration | Joint1 | Joint2 | Joint3 | Joint4 | Joint5 | Joint6 |
|---|---|---|---|---|---|---|
| Right-Down-Fold | $q_{1,1}$ | $q_{2,1}$ | $q_{3,1}$ | $q_{4,1}$ | $q_{5,1}$ | $q_{6,1}$ |
| Right-Up- Fold | $q_{1,1}$ | $q_{2,2}$ | $q_{3,2}$ | $q_{4,2}$ | $q_{5,1}$ | $q_{6,1}$ |
| Left-Down- Fold | $q_{1,2}$ | $q_{2,3}$ | $q_{3,3}$ | $q_{4,3}$ | $q_{5,2}$ | $q_{6,2}$ |
| Left-Up- Fold | $q_{1,2}$ | $q_{2,4}$ | $q_{3,4}$ | $q_{4,4}$ | $q_{5,2}$ | $q_{6,2}$ |
| Right-Down-Unfold | $q_{1,1}$ | $q_{2,5}$ | $q_{3,5}$ | $q_{4,5}$ | $q_{5,3}$ | $q_{6,3}$ |
| Right-Up- Unfold | $q_{1,1}$ | $q_{2,6}$ | $q_{3,6}$ | $q_{4,6}$ | $q_{5,3}$ | $q_{6,3}$ |
| Left-Down- Unfold | $q_{1,2}$ | $q_{2,7}$ | $q_{3,7}$ | $q_{4,7}$ | $q_{5,4}$ | $q_{6,4}$ |
| Left-Up- Unfold | $q_{1,2}$ | $q_{2,8}$ | $q_{3,8}$ | $q_{4,8}$ | $q_{5,4}$ | $q_{6,4}$ |

## 4.2 Numerical Solution

Numerical method is an alternative way for finding the inverse kinematics, The Gauss - Newton iterative method can be used due to the nonlinearity of the model [17]. The nonlinearity caused due to the presence of the sines and cosines in the model. For implementing the Gauss - Newton iterative method, the Jacobian matrix need to be calculated. The Jacobian matrix is ($6 \times n$) matrix consist of the two parts; 1st is the linear Jacobian $J_v$ ($3 \times n$) matrix, and the 2nd is the angular Jacobian ( $3 \times n$) matrix, where n is the robot DOF, in our casen = 6.

$$J = \begin{bmatrix} J_v \\ J_w \end{bmatrix}; \text{ where } J_v = \begin{bmatrix} \frac{\delta P_x}{\delta q_1} & \frac{\delta P_x}{\delta q_2} & \cdots & \frac{\delta P_x}{\delta q_6} \\ \frac{\delta P_y}{\delta q_1} & \frac{\delta P_y}{\delta q_2} & \cdots & \frac{\delta P_y}{\delta q_6} \\ \frac{\delta P_z}{\delta q_1} & \frac{\delta P_z}{\delta q_2} & \cdots & \frac{\delta P_z}{\delta q_6} \end{bmatrix} \text{ and } J_w = \begin{bmatrix} z_1^0 & z_2^0 & \cdots & z_6^0 \end{bmatrix} \qquad (39)$$

$Z_i^0(i = 1 \rightarrow 6)$ Is the 3rd column of the transformation matrix $A_i^0$ , which represents the Z axes of the rotation matrix. The differential kinematics is represented in this form

$$\dot{X} = J\dot{q} \qquad (40)$$

Where $\dot{X}$ is a 6x1 vector of the linear and angular velocities, and $\dot{q}$ is a 6x1 vector for the joints velocities. For a time instance

$$\Delta X = J.\Delta q$$
$$X_d - X_0 = J(q_d - q_0) \qquad (41)$$

Where $X_d$ and $X_0$ are the desired and initial position respectively. $q_d$ and $q_0$ are the desired and initial joint positions respectively. Then the desired velocity can be estimated by inverting the Jacobian matrix in this way

$$q_d = q_0 + J^{-1}\Delta X \qquad (42)$$

So

$$q_d = q_0 + J^{-1}\begin{bmatrix} \Delta X_v \\ \Delta X_R \end{bmatrix} \qquad (43)$$

Where $\Delta X_v$ is $3 \times 1$ vector of the end effector linear velocity. $\Delta X_R$ Is $3 \times 1$ vector representing the angular velocity $\omega$ in $x, y, z$ directions. $\Delta X_v$ is the difference between the desired position $P_d$ and the current initial position $P_0$ in $x, y, z$.

$$\Delta X_v = \begin{bmatrix} P_{xd,yd,zd} - P_{x0,y0,z0} \end{bmatrix} \qquad (44)$$

And the angular velocity $\Delta X_R$ is represented as

$$\Delta X_R = \dot{R}.R_0^T = s(\omega_{x,y,z}) \qquad (45)$$

Where $\dot{R}$ is the derivative of the rotation matrix, $R_0$ is the initial orientation, and S $(\cdot)$ is a skew-symmetric matrix that, for the 3-dimensional case, has the form

$$s(\omega) = \begin{bmatrix} 0 & -\omega_z & \omega_y \\ \omega_z & 0 & -\omega_x \\ -\omega_y & \omega_x & 0 \end{bmatrix} \qquad (46)$$

By Combining the position and orientation we get the equation for the inverse deferential

kinematics as follow

$$q_d = q_0 + J^+ * \begin{bmatrix} P_{xd} - P_{x0} \\ P_{yd} - P_{y0} \\ P_{zd} - P_{z0} \\ \omega_x \\ \omega_y \\ \omega_z \end{bmatrix}$$ (47)

Where $J^+$ is the pseudo inverse of the Jacobian. As In some cases due to singularity it is not possible to invert the Jacobian matrix; therefore we need to use a numerical approximation method such as pseudo inverse. In case if J matrix is full rank, the pseudo inverse will be same as the inverse $J^+ = J^{-1}$, otherwise it will be calculated as $J^+ = J^T (JJ^T)^{-1}$.

For now The Gauss - Newton iterative method in equation (47) is ready for implementation; it required an initial guess for the joints position. The good initial guess usually has good convergence properties [18].

## 5. METHODS EVALUATION

The closed form solution is an optimum solution for the inverse kinematics, in case if it's obtainable. The 8 solution of the UR robot shown in Figure6, for the shoulder, elbow, and wrist configurations, at end effector pose $^0_6T_{desired}$.

$$^0_6T_{desired} = \begin{bmatrix} -0.8421 & -0.4673 & 0.2693 & 0.1942 \\ -0.5196 & 0.5690 & -0.6374 & -0.3593 \\ 0.1446 & -0.6766 & -0.7220 & 0.1701 \\ 0 & 0 & 0 & 1 \end{bmatrix}$$

The axis angle representation for $^0_6T_{desired}$. is $X_{desired}$= {0.1942, -0.3593,  0.1701, -0.8567,  2.7197, -1.1407}.  The 8 solutions for the joints at this pose are shown in

Table 3.

However there are some restrictions should be handled due to singularities. For instance, the robot has singular value if $S_5 = 0$ at ($q_5 = 0, q_5 = \pi$), therefore $q_6$ in equation (11) is not defined, and we can get $q_6$ just when (S5 ≠ 0). Also the physical limits for the robot should be considered, especially when getting the values for $q_2$, $q_3$ and $q_4$, and to be sure that the desired end effector pose within the robot workspace. In the other side, the numerical method can provide only one solution, with a surprising configuration from one of the 8 solutions. We can control over the output of the numerical method by setting arbitrarily the initial conditions. For instance, if the desired configuration is elbow (up) then the initial condition for $q_2$ can be set in that range. Similarly for (left and right) shoulder the $q_1$ can have initial condition in the desired range.

From the point of computational time, the closed form solution for sure can fulfil the real time control requirement. The numerical method is much slower; enhancing its computational time is mainly depends on the initial conditions. Notice that if the initial condition for the numerical method is far away from the desired pose, then it could take long time or it might not converge at all. So when we evaluated the computational time for both methods, we just considered the cases which have good estimate of the initial condition for the numerical method. The computational time to get the inverse kinematics of a single point for the numerical method is 30~50ms, while for the closed form solution is 1~2ms. The difference is very obvious between the two methods in term of computational time.

For evaluating the accuracy of both methods, we considered the Max Absolute Error (MAE) between the desired position and the results of IK of both methods solutions. The pose (position + orientation) MAE of the numerical method is about ( $0.6 * 1e^{-7}$ ), and for the closed form solution about ( $0.2 * 1e^{-7}$ ). The closed form solution absolutely more accurate, since the numerical method is an approximation method; however both methods are acceptable regarding to the limitations of each of them. The IK closed form helped us in finding several robot configurations, thus for adequate workspace in our puncturing robotics system. It can be seen from figure that each configuration gives more space in certain locations in the workspace; demonstrated by the shaded shapes. Doctors can choose the most suitable configuration according to their preferences during the operation. Doctors heights might vary from one to another, so elbow (U-D) configurations could be very helpful in that case. In case if doctors prefer to stand at the left or the right side of the puncture insertion point, the shoulder (R-L) configuration could be very helpful in that case. Anyhow the best configuration can differ from one doctor to another, so this option in our system could make the coexistence of human and robot in the same workspace more adequate.

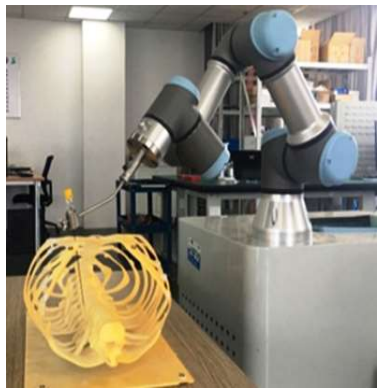 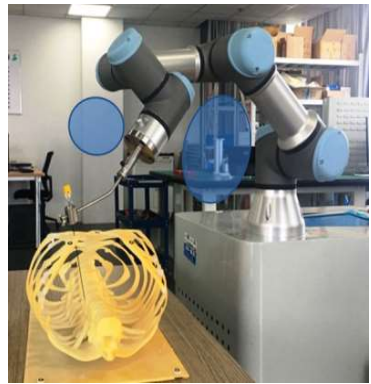

a)  Left- Up- Unfold              b)   Left- Up- Folded

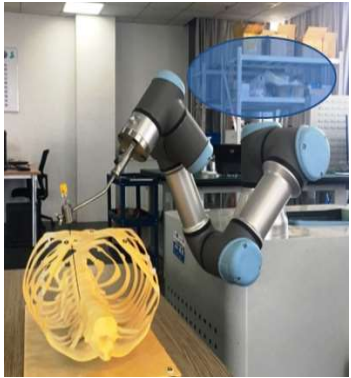
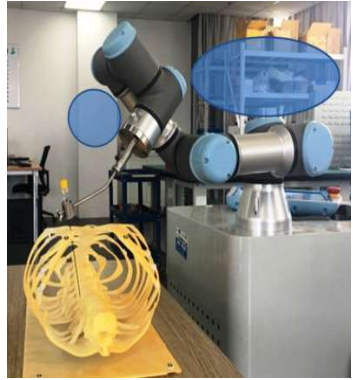

c) Left- Down- Unfold        d) Left- Down- Folded

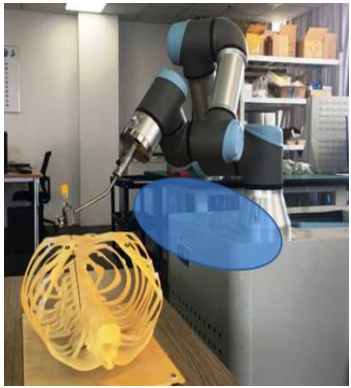
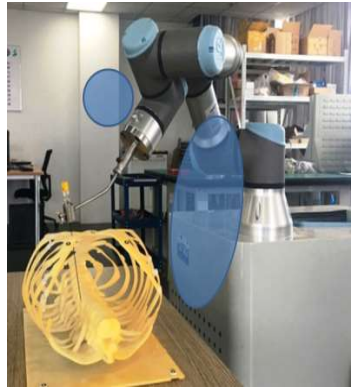

e) Right- Up- Unfold        f) Right- Up- Folded

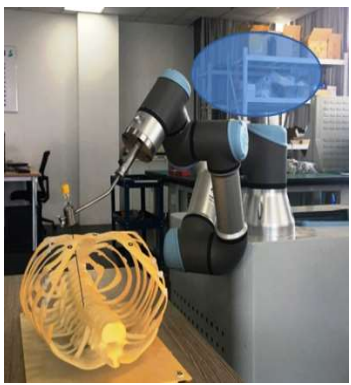
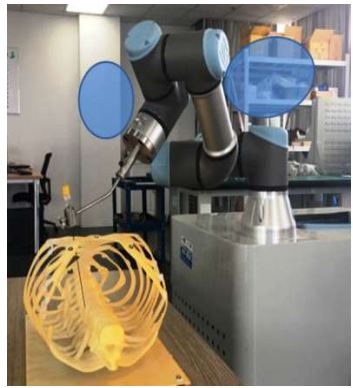

g) Right- Down- Unfold        h) Right- Down- Folded

Figure**6.** UR robot IK solutions, the shaded shapes demonstrate the free space for each configuration

Table 3. TheUR3 inverse kinematics for X= {0.1942, -0.3593,   0.1701, -0.8567,   2.7197, -1.1407}

| Configuration | Joint1 | Joint2 | Joint3 | Joint4 | Joint5 | Joint6 |
|---|---|---|---|---|---|---|
| Right-Down-Fold | -0.73508 | 2.370465 | 1.882823 | -1.96721 | 1.274286 | 0.113862 |
| Right-Up- Fold | -0.73508 | -2.21214 | -1.88282 | 6.381036 | 1.274286 | 0.113862 |
| Left-Down- Fold | -0.73508 | 3.086748 | 0.813237 | -4.7555 | -1.27429 | -3.02773 |
| Left-Up- Fold | -0.73508 | -2.44049 | -0.81324 | 2.398207 | -1.27429 | -3.02773 |
| Right-Down-Unfold | 1.756916 | 0.776095 | -1.87212 | -1.22744 | -1.42357 | -0.36595 |
| Right-Up- Unfold | 1.756916 | 5.367374 | 1.872123 | -9.56297 | -1.42357 | -0.36595 |
| Left-Down- Unfold | 1.756916 | 0.054278 | -0.82719 | 1.591031 | 1.423566 | 2.775645 |
| Left-Up- Unfold | 1.756916 | 5.568666 | 0.827187 | -5.57773 | 1.423566 | 2.775645 |

## 6. CONCLUSION

In this paper we presented the derivation steps of the IK using a combination of analytical and geometric techniques. Closed form IK ensured the most important characteristics of real-time and accuracy of the robot. The innovative application of this work is used in the precise positioning of medical puncture surgery. We showed that how utilizing a robot with obtainable closed form solution IK, such as UR, in puncture robotics system can lead for more adequate workspace. The multiple solutions of IK allow us to develop the system, so that we can offer doctors more robot configurations during the operation.  The selection of the configuration might be according to the one best fit the doctors physically, gives more space in certain direction, gives more visibility, etc. the best configuration can differ from one doctor to another, so this option in our system makes the coexistence of human and robot in the same workspace more adequate.

## AUTHORS

**Omar Abdelaziz persued** his PhD in University of science and technology of China. He is associate lecturer in Egyptian Russian University, Egypt. And Graduate researcher at Institute of intelligent Manufacturing Technology, Jiangsu industrial and technology research institute, China. In 2011 He had received master degree of mechanical engineering from UTHM, Malaysia. Omar's research interest in robotics and intelligent technology. Including topics such as human robots collaboration, serial robots kinematics and dynamics, and Robot control.

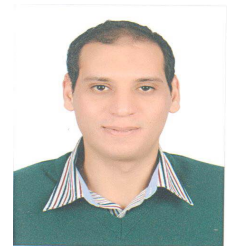

**Minzhou Luo** had received his M.S. degree in Mechanical Engineering from Hefei University of Technology in 2002. He was awarded his Ph.D. degree from University of Science and Technology in 2005.he achieved the researcher of Hefei Institute of Physical Science, Chinese Academy of Science. His research direction is the humanoid robot, industrial robot, Mechatronics, intelligent machine and so on.

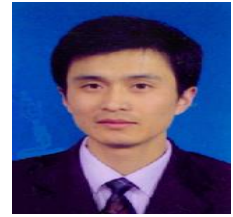

**Guanwu Jiang** is a Ph.D. candidate at University of Science and Technology of China since September 2015. He received the master degree at Southwest University Science and Technology. He studied the kinematics technology and medical image processing technology of medical robots.

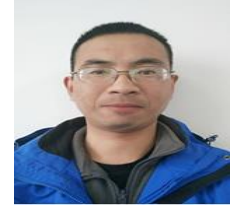

**Saixuan Chen** is a Ph.D. candidate at University of Science and Technology of China since September 2015. He received the bachelor and master degree at Changzhou University. His research interests include the humanoid robot, the structure design of the flexible mechanism, kinematic control of robots, and algorithms for collision detection/reaction.

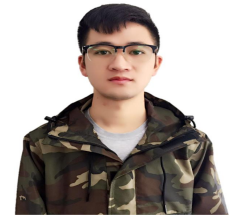